\documentclass[sigconf,nonacm]{acmart}

\usepackage[english]{babel}
\usepackage{blindtext}
\usepackage{tikz}
\usepackage{pifont}
\usepackage{algpseudocode}
\usepackage{algorithm}
\usepackage{multirow}
\usepackage{svg}
\usepackage[font=small,format=plain,labelfont=bf,textfont=bf]{caption}
\usepackage{enumerate}
\usepackage{paralist}
\usepackage{enumitem}[noitemsep,topsep=0pt,leftmargin=0pt]
\setitemize{noitemsep,topsep=0pt,parsep=0pt,partopsep=0pt}
\setenumerate{noitemsep,topsep=0pt,parsep=0pt,partopsep=0pt}
\usepackage[capitalize]{cleveref}
\crefname{section}{\S}{\S\S}

\renewcommand\footnotetextcopyrightpermission[1]{} 
\setcopyright{none}

\acmDOI{}

\acmISBN{}

\acmConference[Submitted for review to APSys]{}
\acmYear{2024}

\acmPrice{}

\newif\ifsubmission
\submissiontrue

\ifsubmission
\newcommand{\mcnote}[1]{}
\newcommand{\todo}[1]{}
\newcommand{\banruo}[1]{}
\newcommand{\mubarak}[1]{}
\else
\usepackage[textsize=tiny]{todonotes}
\newcommand{\mcnote}[1]{\todo[color=orange!40,inline]{marco: #1}} 
\newcommand{\banruo}[1]{{\color{blue}[Banruo: #1]}}
\newcommand{\mubarak}[1]{{\color{yellow}[Mubarak: #1]}}
\fi
\newcommand{\sys}{NeuronaBox\xspace}
\newcommand{\smartparagraph}[1]{\noindent{\bf #1}\ }

\def\nbReal{{\mathcal{N}}}
\def\nbEnv{{\mathcal{E}}}

\begin{document}

\title[Towards a Flexible and High-Fidelity Approach to Distributed DNN Training Emulation]{Towards a Flexible and High-Fidelity Approach to\\ Distributed DNN Training Emulation}

\author{Banruo Liu}
\orcid{0009-0002-9932-6096}
\authornote{Work done primarily while author was interning at KAUST.}
\affiliation{%
  \institution{Tsinghua University}
  \country{}
}

\author{Mubarak Adetunji Ojewale}
\orcid{0000-0003-3861-1782}
\affiliation{%
  \institution{KAUST}
  \country{}
}

\author{Yuhan Ding}
\orcid{0009-0000-2829-0319}
\affiliation{%
  \institution{Tsinghua University}
  \country{}
}

\author{Marco Canini}
\orcid{0000-0002-5051-4283}
\affiliation{%
  \institution{KAUST}
  \country{}
}

\begin{abstract}

We propose \sys, a flexible, user-friendly, and high-fidelity approach to emulate DNN training workloads. We argue that to accurately observe performance, it is possible to execute the training workload on a subset of real nodes and emulate the networked execution environment along with the collective communication operations. Initial results from a proof-of-concept implementation show that \sys replicates the behavior of actual systems with high accuracy, with an error margin of less than 1\% between the emulated measurements and the real system.

\end{abstract}

\maketitle

\section{Introduction}
\label{sec:introduction}

Modern DNN training clusters are remarkable engineering feats that more closely resemble high-performance specialized computing environments -- and the large costs that these entail -- than their mainstream counterparts in commodity cloud computing datacenters.
Optimizing resource utilization and overall efficiency is paramount to maximizing the performance of training workloads and minimizing associated costs.
Therefore, it is highly desirable to explore the large space of potential design considerations, performance optimizations and configuration tunings, and ideally to do so without incurring time, energy, and monetary costs of profiling training workloads at scale on actual hardware.

Conducting in-depth ``what if'' analyses is essential to making informed decisions and beneficial for a variety of scenarios. For instance, a ML engineer may want to explore for a given model the impact of a particular parallelization strategy\footnote{Possible strategies include data parallelism~\cite{xing2015petuum}, tensor parallelism~\cite{shoeybi2020megatronlm}, pipeline parallelism~\cite{huang2019gpipe}, fully sharded data parallelism~\cite{zhao2023pytorch} among others.} on the training time and resource utilization.
But it is not practical to profile the training workload on thousands of HW accelerators (GPUs, TPUs, etc.) for each possible strategy and different configurations.
Similarly, a researcher may want to quantify the benefits \emph{at scale} of a new optimization technique to improve LLM training efficiency. Also in this case, it is hardly feasible to run systematic experiments on a large cluster for each possible configuration.

Recent work has shown the potential of simulation and analytical methods to gain insights about DNN training behavior~\cite{rashidi2020astra, rashidi2022themis, esposito2022dts, ardalani2022deepflow, won2023astra, bang2023vtrain}. However, these approaches suffer from at least one of three limitations:
\begin{inparaenum}[1)]
\item they require significant effort to transform the actual workloads into an input model for the simulator,
\item they require explicit models of parallelization strategies and incorporating new ones entails non-trivial development of new simulation models, and,
\item the fidelity of their results is limited by how faithful the underlying analytical models of compute and communication are, which are notoriously difficult to get right at scale~\cite{mittal2018rocerdma}.
\end{inparaenum}

\begin{figure}[t!]
    \centering
    \includegraphics[trim={0 1.2cm 0 0},clip, width=0.975\linewidth]{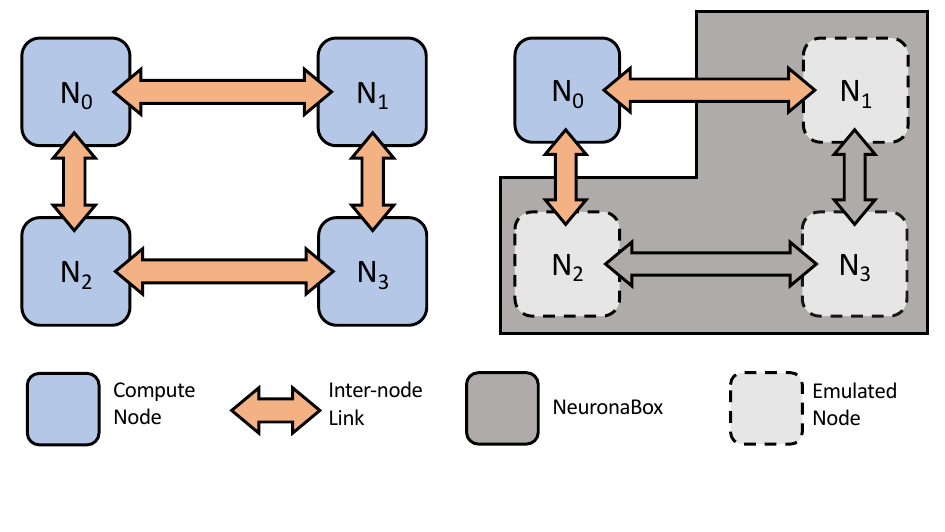}

\caption{A training job running in a 4-node cluster (left) is emulated by executing a single real node ($N_0$) wrapped by \sys, which emulates the environment (right).}
    \label{fig:neuornabox}
\end{figure}

This work pioneers and advocates the use of emulation to aid in the analysis and experimentation of distributed DNN training workloads.
What we mean by emulation is illustrated in \cref{fig:neuornabox}. In a nutshell, we propose to isolate a node subset (denoted as $\nbReal$) of a distributed training job and emulate the networked execution environment (denoted as $\nbEnv$) from the perspective of the nodes in $\nbReal$.
We elect to view the network as a natural boundary between the real and emulated environments since communication between nodes in distributed training jobs typically occurs through a collective communication library (e.g., NCCL~\cite{nccl}) that both isolates the training scripts from dealing with all the unnecessary details of the underlying network and demarcates clear points for inter-process synchronization.
We refer to our approach as \sys.

Notably, in this approach, the nodes in $\nbReal$ run unmodified training scripts, DNN frameworks and libraries. In particular, the communication is handled by the actual collective communication library over the network fabric.
Meanwhile, the emulation environment $\nbEnv$ executes on dedicated hardware resources. The requirements for the emulation environment are modest: it doesn't require HW accelerators, it can run on a single CPU-based node, and it requires network bandwidth to match the available aggregate bandwidth of nodes in $\nbReal$.

The key benefit of this approach is that it allows us to faithfully execute on real hardware a portion of the training workload, which executes without overheads from instrumentation (since there is none) nor profiling $\nbReal$ in controlled conditions.
Therefore, we can observe the actual behavior of the training job, including the HW utilization metrics and collective communication patterns that are critical in analyzing the performance of distributed training workloads.

We wish to stress that our objective is to enable performance analysis and optimization of distributed training workloads. Implications on model quality are out of scope.

Thus, in this work, we initiate the study of these core research questions:
\begin{inparaenum}[1)]
    \item \emph{What aspects of the workload must $\nbEnv$ emulate?}
    \item \emph{How can this approach maintain high fidelity while retaining wide applicability?}
\end{inparaenum}

With this short paper, we aim to provide an initial exploration of the feasibility and potential of this approach, and to solicit feedback from the community on the soundness of our approach and the prospects for future research.

\section{Proposed Approach}
\label{sec:design}
Our goal is to enable any subset $\nbReal$ of nodes in a distributed DNN training job to execute the workload as if it were running on the entire set of nodes and resources. We propose to achieve this goal by emulating the interactions between $\nbReal$ and its networked environment $\nbEnv$, which in a sense can be viewed as a virtualization of the remaining job's nodes.
We argue that, under certain assumptions (detailed below), by observing the performance of $\nbReal$, we can analyze and extrapolate the behavior for an entire job with high fidelity.

In our design, we adhere to two driving principles:

\smartparagraph{1) Ease of use.}
The user should be able to use \sys without any modification to their existing code.

\smartparagraph{2) Flexibility and independence of parallelization strategies.}
\sys should target a level of abstraction that is independent of the specific parallelization strategy.
\sys should be flexible to seamlessly adapt to changes in parallelization strategies, including new ones that may emerge in the future.

\begin{figure}[t!]
  \centering
  {
\includegraphics[width=0.975\linewidth]{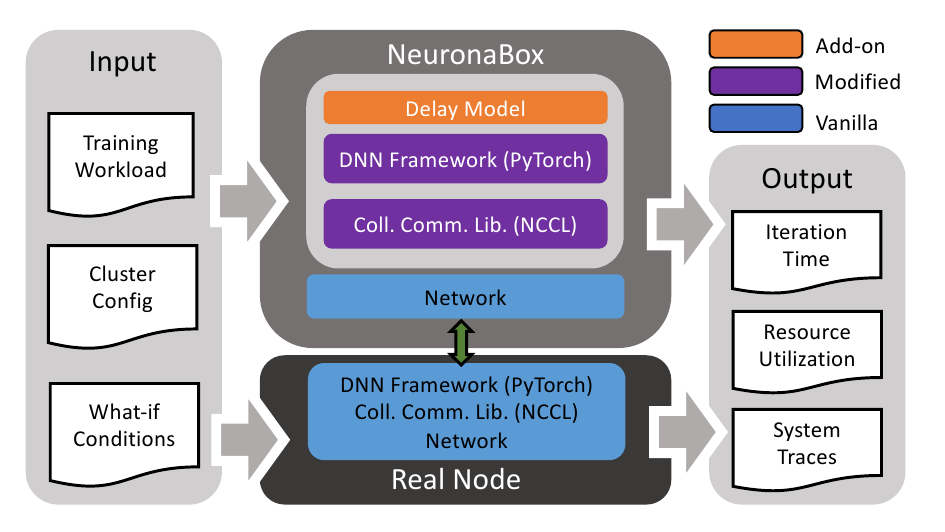}
  }
  \caption{Overall workflow and architecture of \sys.}
  \label{fig:overview}
\end{figure}

\smartparagraph{Workflow and architecture.}
\cref{fig:overview} depicts an overview of our approach.
The high-level workflow of \sys is as follows.
First, the user provides the training script, the job configuration (e.g., world size, nodes in $\nbReal$, HW resources, etc.), and optionally a set of what-if conditions for experimentation (an example is given later).
Second, \sys initializes the emulation environment by synthesizing the network topology and instantiating a communication model that calculates delay times for collective operations within the emulated environment.
Third, the training script is launched (e.g., via \textsf{torchrun}). Meanwhile, desired performance metrics like iteration time and resource utilization are gathered in $\nbReal$. Traces of collective communication (e.g., NCCL traces) can also be collected.

\smartparagraph{Assumptions.}
We assume that nodes have uniform hardware and network configuration.
In practice, it is common to execute distributed training jobs on homogeneous clusters ~\cite{wang2023zero++,osdi22alpa,touvron2023llama,liu2019roberta,lai2023adaembed}. 
We assume that the model fits entirely within $\nbReal$. This assumption is not restrictive, as it is common to use model or tensor parallelism within a node or a shard~\cite{hwang2023tutel,shoeybi2020megatronlm} . 
We expect that these assumption yield a sort of symmetry in the workload distribution across the nodes, which allows us to treat the nodes in $\nbReal$ as a representative sample of the entire nodes.
We discuss how to extend our arguments to a non-uniform scenario in \cref{sec:design:non-uniform}.

Further, we assume that the collective communication layer is the only point of interaction between $\nbReal$ and $\nbEnv$. This assumption is reasonable, as the collective communication layer is the primary interface between the computation and the network stack in distributed training jobs.
Finally, note that we are free to modify the DNN framework and collective communication libraries within the emulator. That is how we are able to implement \sys!

\smartparagraph{Scalability.}
While collective communication is a natural layer to target in our work, the astute reader may now wonder how scalable this approach is.
Scalability is traditionally a challenge in network emulators~\cite{mininet,modelnet}, as emulating a large number of nodes could overload the emulator.
Our key insight is that we are only interested in the interaction between $\nbReal$ and the outside world. And so, the actual communication between the emulated nodes can be skipped.
Instead, only the delay resulting from these communication operations needs to be incorporated into the emulation. As a result, the number of connections as well as the amount of data transfer for \sys are the same as that of $\nbReal$. This observation allow \sys to potentially scale to a large number of nodes. A complete exploration of the scalability of \sys is left for future work.

In the remainder of this section, we detail the design and implementation of our proposed approach, \sys. Since we aim to demonstrate feasibility through a proof-of-concept, we focus on a single real node, denoted by $N_0$. That is, $\nbReal = \{N_0\}$.

\subsection{Initialization} \label{sec:design:init}

Before we describe how \sys behaves during training, we first need to initialize the emulator. This requires setting up the environment, including the network topology and the communication model.

\smartparagraph{Topology detection.}
\label{sec:design:topology}
This step involves establishing the connection between any pair of nodes, and finding the optimal data paths between all node pairs. It is normally bootstrapped by the collective communication library itself. During this process, each node builds up its own local topology graph (how HW accelerators are connected via NVLink, PCIe and NICs), then exchanges its local graph and, together with other nodes, builds the global graph.
In \sys, the emulator fakes the local graph of emulated nodes based on the job configuration input. Then, it emulates multiple end points (the virtualized job nodes) so that nodes in $\nbReal$ can communicate with them (directly via RDMA).

\smartparagraph{Delay calculation.}
Based on the global topology, \sys calculates the communication and computation delays within the emulated environment. In our design, we provide an interface so that the calculation itself is done within a user-defined add-on plugin. For example, the delay in a ring all-reduce call can be estimated simply by using the classical all-reduce delay model~\cite{thakur2005optimization}, or by using a packet-level simulator.
In the future we seek to leverage the network simulation components of DNN simulators~\cite{ardalani2022deepflow, won2023astra, bang2023vtrain}.
\subsection{Emulation in a Uniform Scenario} \label{sec:design:uniform}

\begin{figure}[t!]
  \centering
{
\includegraphics[trim={0 0.7cm 0 0},width=0.975\linewidth]{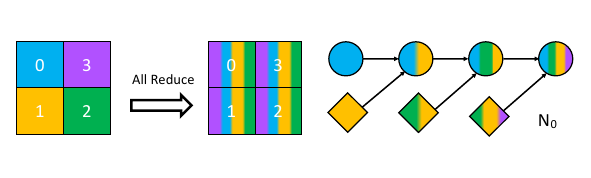}
\includegraphics[width=0.975\linewidth]{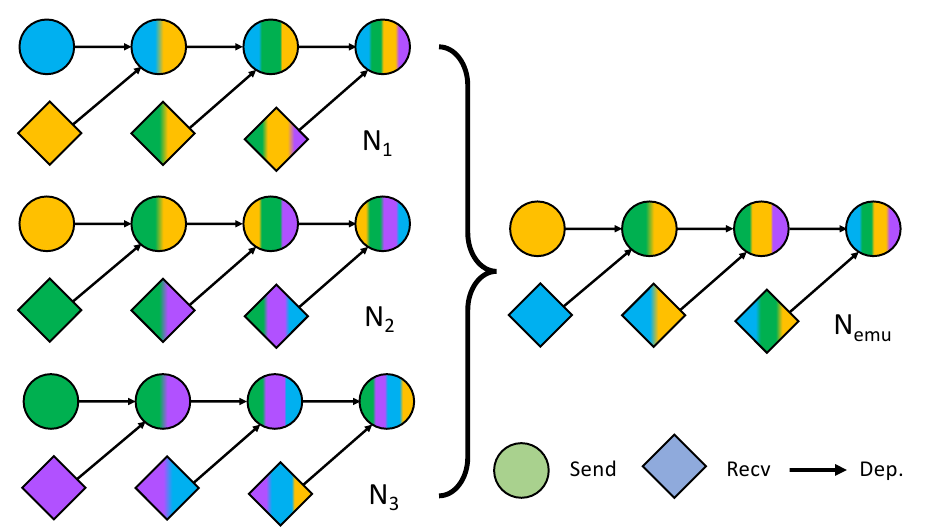}
  }
  \caption{An example DAG for four-node ring all-reduce. The upper left squares shows the net result of all-reduce, where color-coded data from different nodes are reduced and then gathered at each node. The upper right figure shows the dependency DAG for $\nbReal$ ($N_0$). The lower figure shows how we merge the dependency DAG of $N_1, N_2, N_3$ into $\nbEnv$. The cross-node dependencies from $send(x)$ to $recv(x)$ are not shown for clarity sake. We only show the initial 4 steps of all-reduce for simplicity. }
  \label{fig:dag}
\end{figure}

In general, a collective operation (e.g., ring all-reduce) can be split into a number of messages with dependencies. The emulator must send/receive messages in a way that takes into account both the dependency and the internal protocol of the collective operations library (in our case, NCCL). We first describe how we treat a single collective operation and then generalize to multiple asynchronous collective operations.

\smartparagraph{Single collective operation.}
Recall that we only need to consider the interaction between $\nbReal$ and $\nbEnv$. This means that we can omit the dependencies for communication within $\nbEnv$.
The workflow of a collective operation can be represented as a directed acyclic graph (DAG) where vertices are send or recv tasks and edges are the data dependencies.
Thus, we note that in \sys, this DAG can be greatly simplified. \cref{fig:dag} shows an example for ring all-reduce.\footnote{The all-reduce operation performs reduction on data (i.e., sums) across nodes and stores the result in a buffer at every node. We use different color for data in different worker.} It is worth noting that the DAGs for $\nbReal$ and \sys are isomorphic. As a result, with every message received from $\nbReal$ to $\nbEnv$, we can always determine the correct state in order to generate the next message of the collective operation workflow.

To achieve this, $\nbEnv$ maintains a bitmap of the messages that have been sent to or received from $\nbReal$, and it applies the following two actions ($\nbEnv$ polls these using background threads) that advance the state of the DAG ensuring synchronization correctness:

\noindent 1) \textit{Try Send To $\nbReal$}. A message can be sent if and only if all its predecessors have been sent or received in the DAG. If the next unsent message in the bitmap fulfils this condition, we update the bitmap and send the message.

\noindent 2) \textit{Try Receive From $\nbReal$}. Upon receiving a message from $\nbReal$, $\nbEnv$ checks whether it is the expected message for the current operation. If this is the case, the bitmap of the record is updated; otherwise, an error is reported.

\smartparagraph{Multiple async collective operations.}
The design of a single collective operation can easily be extended to support multiple operations by assigning a unique ID to each operation and maintaining its information in a logically centralized controller. We record the mapping of operations to their message bitmaps. We ensure fairness between the individual streams through round-robin polling. Note that the method for single operation is asynchronous by nature, as the functions are polled by background threads. Given that, the synchronization is achieved by busy-waiting. 

\subsection{Extension to a Non-uniform Scenario} \label{sec:design:non-uniform}

The workload of individual nodes in a training job may not always be balanced. This is the case, for example, when model parallelism fails to achieve a balanced workload distribution, or when there is heterogeneity of hardware and topology. Consequently, emulating the behavior of an arbitrary node subset ($\nbReal$) may not adequately represent the behavior of the entire workload. To overcome this challenge, we propose to classify each node in the job based on the part of the model it contains, e.g., like having different stages in model parallelism. We then ensure that one node in each class is in $\nbReal$. This approach allows us to infer the behavior of the workload by observing the collective behavior of each class of nodes. This approach also solves the heterogeneity of hardware and topology when we classify nodes with different hardware into different classes.

However, we note that this approach requires more resources. Suppose there are $m$ classes; in the basic solution, we need to have one representative real node for each of them. And with $t$ hardware types, we then need $tm$ nodes in $\nbReal$.
We conjecture that it may be possible to reduce the number of nodes in $\nbReal$, say to $k$, by decoupling the emulation between different nodes. Assuming nodes in $\nbReal$ are in different classes, if all the communication is interposed by $\nbEnv$, then we can time-multiplex the class-based workload and assign each node $\frac{m}{k}$ amount of load.
We leave the exploration of these techniques for future work.

\begin{table}[t!] 
\resizebox{\columnwidth}{!}{%
\begin{tabular}{|r|r|r|r|r|}
\hline
Size  & AllreduceB & AllreduceE & AllgatherB & AllgatherE \\ \hline
1KB   &      435.2us       &    418.8us         &   282.6us          &      276.2us       \\ \hline
4KB  &        526.5us     &      511.0us      &     306.5us       &       300.2us      \\ \hline
32KB &    564.9us         &  552.3us          &    329.0us        &   322.6us          \\ \hline
256KB &      1326.0us       &        1314.0us    &     868.9us        &   859.6us         \\ \hline
2MB &      7661us      &      7655us      &  4928us           &   4929us          \\ \hline
16MB &        59.0ms     &       58.9ms      &        37.5ms     &      37.5ms       \\ \hline
128MB &     470ms        &     469ms        &    298ms         &      298ms       \\ \hline
1GB &     3760ms        &    3760ms         &  2408ms           &   2407ms          \\ \hline
\end{tabular}
}
\caption{Average run time per call. B for baseline and E for emulator (\sys). \sys only incurs at most 4\%/2\% extra time for all-reduce and all-gather, respectively.}
\label{tab:coll}
\end{table}

\subsection{Proof-of-concept Implementation}

Our proof-of-concept implementation entails the development of an end-to-end system using the PyTorch DNN framework and NCCL as the collective communication library, chosen because of their popularity. 

Our implementation is able to run a two-node training using a distributed data-parallel strategy. In particular, we modify NCCL's instance in $\nbEnv$ to emulate the collective operations as per \cref{sec:design:uniform}. Moreover, in $\nbEnv$ we skip the \textsf{cudaKernelLaunch} completely so that no GPU computations are involved. We also modify the NCCL proxy so that it sends dummy data in compliance with the internal protocol so that the workload continues to run and $\nbReal$ is not aware of the emulation. In PyTorch, we mainly alter the \textsf{autogard} and \textsf{c10d} to implement the synchronization that previously relies on a CUDA kernel now compatible with the emulator.  We also remove computation in backward pass and model weight update in $\nbEnv$, given that those computations are redundant in emulation.

The whole system is about 2000 LoC in CUDA C++ and 50 LoC in Python, exclusive of experiments and tests.
We plan to release \sys as open source.

\section{Preliminary Experiments}
\label{sec:experiments}

We evaluate our prototype \sys implementation by ($1$) running microbenchmarks in NCCL level to see the pure performance of collective communication, ($2$) running an end-to-end system in PyTorch to see the accuracy of emulation, as well as measuring CPU utilization to evaluate the overheads of \sys. We also demonstrate ($3$) an application scenario of \sys by performing a ``what-if'' analysis with latency variations.

\vspace{0.1in}
\smartparagraph{Testbed.} Our test environment consists of two nodes, each equipped with two 8-core Intel Xeon Silver 4112 CPUs running at 2.60 GHz, 512 GB RAM and is fitted with a 100 GbE Mellanox ConnectX-5 NIC. In addition, each node contains two NVIDIA V100 GPUs, although only one is used during evaluation. Each node runs Ubuntu 22.04 (Linux kernel 5.15.0), CUDA 12.2, PyTorch 2.2.0a0 and NCCL 2.19.4.

If not otherwise stated, we call the two nodes $\nbReal=\{N_0\}$ and $\mathcal{E} = \{N_{emu}\}$. $N_0$ always runs the unmodified code. $N_{emu}$ is configured to run either \sys's modified code(as emulator), or unmodified modified code (as a baseline).

\subsection{Microbenchmark}

\smartparagraph{Setup.} First we assess \sys's capability to emulate collective communication operations. We devise the benchmark by generating input data tensors of different sizes on GPU and issuing two-node collective operations. We test \textsf{ncclAllreduce} and \textsf{ncclAllgather}. After warm-up, we measure the time taken over at least 100 repetitions for each call on $N_{emu}$ and report the average. $N_0$ always runs unmodified code. We compare the result when $N_{emu}$ is running unmodified NCCL (baseline) and \sys's NCCL (emulator).

\smartparagraph{Results.} \cref{tab:coll} shows that \sys only incurs at most 4\% overhead; we attribute this to the mutex lock on the controller and bitmap bookkeeping. The overhead diminishes as the size increases; when the size is greater than 2MB, the gap is no more than 1\%. Since most data parallel implementations use buckets to batch all-reduce calls to a larger size (e.g., 25MB in NVIDIA APEX ~\cite{nvidiaapex}), we believe this NCCL-level overhead of \sys is acceptable. 

\subsection{End-to-end Training Emulation}

\smartparagraph{Setup.} To evaluate \sys's ability to accurately emulate end-to-end DNN training, we conducted experiments using three real-world DNN models, including computer vision, natural language processing and recommendation systems. The models details are listed in \cref{table:model}. We use PyTorch's \textsf{DistributedDataParallel} module for data parallelism. $N_0$ runs unmodified PyTorch; $N_{emu}$ runs \sys's PyTorch as emulator and unmodified PyTorch as baseline. 

We measure the metrics and report the average after warm-up. We report training time per iteration for BERT and ResNet; we report training time per epoch for DeepLight because it involves model pruning and has variance in between iterations. We also measure the CPU usage in \sys to illustrate the overhead of the emulation.

\begin{table}[t!]
\centering
\resizebox{\columnwidth}{!}{%
\begin{tabular}{||l r r r||} 
 \hline
 Model & Task & Dataset & Size \\ [0.5ex] 
 \hline
 BERT ~\cite{devlin2018bert}& Question Answer  & SQuAD ~\cite{rajpurkar2018squad} &   1.28GB \\ 
 ResNet152 ~\cite{he2016resnet} & Image Classify & ImageNet-1K ~\cite{russakovsky2015imagenet}& 230MB \\
 DeepLight ~\cite{deng2021deeplight} & Click Predict &  Criteo 1TB ~\cite{Criteo1t}  & 2.26GB   \\[0.5ex] 
 \hline
\end{tabular}
}
\caption{Characteristics of benchmark DDN models, size represents the total parameter size of a model.}
\label{table:model}

\begin{tabular}{||l r r r r||} 
 \hline
 Model & Time-E & Time-B & CPU-E & CPU-B \\ [0.5ex] 
 \hline
 BERT & 629 $\pm$ 3.0  & 628$\pm$1.1 & 12.93\% & 14.25\%  \\ 
 ResNet152 & 1061$\pm$19.8 & 1063$\pm$16.3  &12.68\% &12.95\% \\
 DeepLight & 727$\pm$15.0  & 726$\pm$ 13.8  & 7.52\% & 7.75\%   \\[0.5ex] 
 \hline
\end{tabular}

\caption{End-to-end workload comparison. `E' and `B' stand for emulator-enabled (\sys) and the baseline, respectively. `Time' stands for the training times in milliseconds; and `CPU' stands for the percentage of CPU usage in a node.}
\label{table:e2e_results}
\end{table}

\smartparagraph{Results.} As shown in \cref{table:e2e_results}, \sys is quite accurate in a two-node environment training with data parallelism, with error less than 1\%. 
For CPU usage, it actually drops a little bit in all scenario. We attribute that to: ($1$) the efficient and lightweight implementation of \sys, which keeps the overhead generally low; ($2$) the removal of computation in backward pass, which eliminates a lot of memory allocation and data movements. So the net effect is a drop in CPU usage. This is promising in terms of the potential scalability of \sys.

\subsection{What-if Analysis: Latency}

\smartparagraph{Setup.} In this experiment, $N_0$ runs unmodified code and $N_{emu}$ runs \sys. We inject additional delay in each all-reduce call in $N_{emu}$ and train a BERT model. We measure the emulated training time per iteration for each delay.

\begin{figure}[t!]
\centering 
\includegraphics[width=0.975\linewidth]{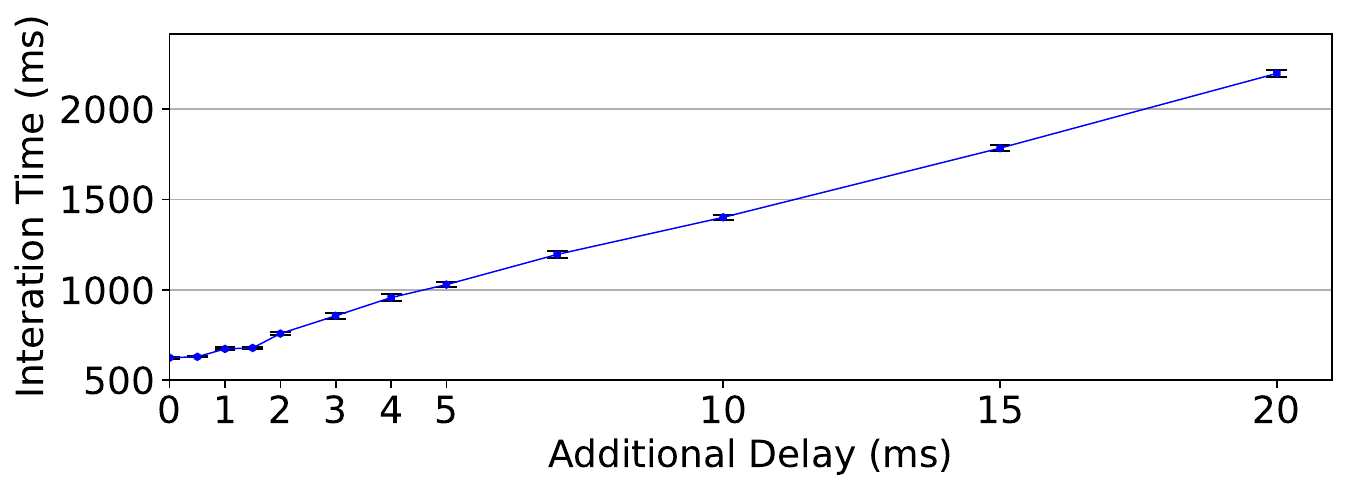}
 \caption{The end-to-end training time per iteration in BERT model (ms) vs the additional delay injected in every all reduce call (ms). Error bar is plotted in black.}
  \vspace{4mm}
  \label{fig:delay}
\end{figure}

\smartparagraph{Results.} As \cref{fig:delay} shows, the iteration time increases linearly with the delay larger than 2 milliseconds. However, when the delay is small, the overheads reflected in the end-to-end performance does not grow linearly. We consider this ``smooth slope'' as a result of the computation-communication overlap during the training. The delay injected in each all-reduce call is partly shadowed by the async computation in the backward pass. Such observation implies a possible space of improvement in the training, as there is a "2ms" space for more communication to happen. This is an example of the kind of insights that can be obtained from using \sys.

\section{Related work}
\label{sec:related_work}
\smartparagraph{DDL simulation.} A number of simulators have been developed to study the behavior of DNN clusters, including DeepFlow\cite{ardalani2022deepflow}, Astra-sim~\cite{rashidi2020astra,won2023astra}, vTrain~\cite{bang2023vtrain} and several others~\cite{rashidi2022themis, esposito2022dts, stanic2023languini, hestness2017deep}. These simulators use analytical methods combined with profiling results to make predictions, suffering from limitations mentioned in \cref{sec:introduction}.

\smartparagraph{Network emulation.} Emulation has been widely adopted in networking research~\cite{modelnet,mininet}. MimicNet~\cite{zhang2021mimicnet} is a machine learning based network emulator. It exercises a similar idea by dividing the datacenter into an ``observable'' cluster ($\nbReal$ in our work) and a blackbox ($\mathcal{E}$ in our work) and it applies a machine learning model to fit it. However, MimicNet focuses on how to train a model to better approximate the datacenter network at scale, whereas our work focuses on emulating end-to-end DNN training behaviors. 

\smartparagraph{Goodput prediction.}
Currently, \sys only sends \textit{dummy} data to $\mathcal{N}$ during emulation since it only predicts the completion time for each training iteration. However, lossy training optimization techniques like compression and quantization ~\cite{wang2023zero++, lin2017gradientcompression} require goodput (accuracy) to be taken into account. To support that, \sys needs to communicate \textit{meaningful} data to $\mathcal{N}$ without incurring much overhead. One possible solution is to use a proxy model~\cite{coleman2019selection} that generates data with a similar distribution to the dataset and intermediate results.

\section{Conclusion}
\label{sec:conclusion}
We proposed a novel approach for estimating time-per-iteration in distributed DNN training, focusing on executing only a part of the model along with collective communication operations. To substantiate our proposal, we designed the \sys emulator and implement a proof-of-concept system. Through extensive experimentation, we demonstrated in a two-node setup that \sys achieves high accuracy in predicting training time across a variety of DNN models, with an error margin of less than 1\% compared to actual training runs. Finally, we encourage further research in this direction, recognizing that many questions remain to be explored.

\bibliographystyle{ACM-Reference-Format}
\bibliography{main}

\end{document}